\documentclass[runningheads]{llncs}

\usepackage[mobile]{eccv}

\usepackage{eccvabbrv}

\usepackage{graphicx}
\usepackage{booktabs}
\usepackage{multirow}
\usepackage{graphicx}
\usepackage{bbding}
\usepackage{algorithmicx,algorithm}
\usepackage{algpseudocode}
\usepackage{color}
\usepackage[rightcaption]{sidecap}
\usepackage[colorlinks=true, urlcolor=magenta]{hyperref}

\usepackage[accsupp]{axessibility}  

\newcommand{\bestscore}[1]{\colorbox[rgb]{0.9686,0.8078,0.6274}{#1}}
\newcommand{\secondscore}[1]{\colorbox[rgb]{1,1,0.6509}{#1}}

\usepackage{hyperref}

\usepackage{orcidlink}

\begin{document}

\title{RS-NeRF: Neural Radiance Fields from Rolling Shutter Images}

\titlerunning{RS-NeRF}


\author{Muyao Niu\and
Tong Chen \and
Yifan Zhan \and
Zhuoxiao Li \and \\
Xiang Ji \and
Yinqiang Zheng\inst{\ensuremath{\star}}
}

\authorrunning{M. Niu et al.}

\institute{
The University of Tokyo \\
\email{muyao.niu@gmail.com}, \email{cocottt1023@gmail.com}, \\ \email{zhan-yifan@g.ecc.u-tokyo.ac.jp}, \email{lizhuoxiao@g.ecc.u-tokyo.ac.jp}, \email{xiangji2016@gmail.com}, \email{yqzheng@ai.u-tokyo.ac.jp}
}

\maketitle

\renewcommand{\thefootnote}{\fnsymbol{footnote}}
\footnotetext[1]{Corresponding author}
\renewcommand{\thefootnote}{\arabic{footnote}}

\begin{abstract}
  Neural Radiance Fields (NeRFs) have become increasingly popular because of their impressive ability for novel view synthesis. However, their effectiveness is hindered by the Rolling Shutter (RS) effects commonly found in most camera systems. To solve this, we present RS-NeRF, a method designed to synthesize normal images from novel views using input with RS distortions. This involves a physical model that replicates the image formation process under RS conditions and jointly optimizes NeRF parameters and camera extrinsic for each image row. We further address the inherent shortcomings of the basic RS-NeRF model by delving into the RS characteristics and developing algorithms to enhance its functionality. First, we impose a smoothness regularization to better estimate trajectories and improve the synthesis quality, in line with the camera movement prior. We also identify and address a fundamental flaw in the vanilla RS model by introducing a multi-sampling algorithm. This new approach improves the model's performance by comprehensively exploiting the RGB data across different rows for each intermediate camera pose. Through rigorous experimentation, we demonstrate that RS-NeRF surpasses previous methods in both synthetic and real-world scenarios, proving its ability to correct RS-related distortions effectively. Codes and data available: \url{https://github.com/MyNiuuu/RS-NeRF}
\end{abstract}

\section{Introduction}
\label{sec:intro}

CMOS imaging sensors are commonly employed in a wide range of consumer and industrial products. Many of these sensors utilize a Rolling Shutter (RS) mechanism for image capture because of its cost-effectiveness and portability. In contrast to Global Shutter (GS) cameras, which capture all pixels simultaneously, rolling shutter cameras capture image pixels row by row in a sequential manner. Consequently, if the camera is in motion during image capture, visible distortions can occur. It is widely acknowledged that rolling shutter distortions pose challenges for various downstream computer vision tasks, as indicated by previous research~\cite{albl2019rolling,hedborg2012rolling,kim2017rrd,klingner2013street,saurer2013rolling,saurer2016sparse}. Consequently, the correction of rolling shutter effects has attracted significant attention in the past~\cite{rengarajan2017unrolling,vasu2018occlusion,zhuang2019learning,liu2020deep,zhong2021towards,zhou2022evunroll,zhong2022bringing,fan2023joint}.

In the emerging field of novel view synthesis, technologies such as NeRF~\cite{mildenhall2021nerf} have become increasingly proficient in generating highly realistic renderings. NeRF utilizes a continuous volumetric function that is parameterized by a multilayer perceptron (MLP) to associate 3D locations and 2D view directions with color and density information. It relies on volume rendering techniques to enable differentiable rendering.

Although traditional NeRF models deliver exceptional results when applied to well-calibrated and well-captured images, they encounter difficulties when dealing with RS effects. For example, in scenarios where the camera is in motion during image capture, these RS effects can significantly impact the performance of the model. Given the widespread use of RS cameras in various industrial applications, it is imperative to take into account this phenomenon when reconstructing 3D representations using NeRFs.

One straightforward approach involves a two-stage baseline solution, where the first step corrects the rolling shutter effect in 2D image space, and then the NeRF is trained using these corrected images. While these baseline methods enhance the quality of novel view synthesis in NeRF to some extent by utilizing 2D rolling shutter correction techniques, they can not fully explore the 3D scene's geometry, leading to inaccuracies in correspondences, especially when there is significant camera motion.

In this paper, we introduce RS-NeRF, a robust framework designed to explicitly incorporate the physics-based formation of rolling shutter effects into the rendering process. RS-NeRF has the ability to reconstruct a high-quality NeRF representation from only input data that contains rolling shutter artifacts. To achieve this, we explicitly model the camera's motion trajectory throughout the exposure duration for each frame and combine the RGB rendering outcomes from different camera poses to generate the rolling shutter image. Throughout this procedure, we optimize both the NeRF network and the pose estimation network simultaneously by minimizing the photometric loss.

Building upon the basic RS-NeRF framework, we delve into its shortcomings and enhance its functionality through various refinements. Initially, we impose a camera trajectory smoothness regularization to refine our model's trajectory prediction, by leveraging the sequential nature of camera movement to yield improved results. We further identify a critical shortcoming due to the vanilla RS camera motion model and introduce a multi-sampling technique that significantly enhances RS-NeRF's performance. Specifically, we reveal the issue of the standard camera pose formulation, which assigns distinct poses per row, potentially limiting RS-NeRF's efficiency due to limited training data for each pose. We then make a key observation that multiple poses may share the same 2D RGB observation for the same 3D point during the exposure time, which can drastically increase the available training data. To capitalize on this insight, we introduce and compute the \textbf{PP-ratio}—a measure of camera pose movement in 3D space relative to pixel movement in 2D space, and apply this to broaden our training data by sampling out-of-row pixel points. This approach significantly bolsters our model's performance by tapping into the inherent data augmentation possibilities presented by the RS formation process.

We collect both synthetic and real data to evaluate our proposed RS-NeRF. The experimental results reveal that RS-NeRF outperforms the preceding models and demonstrate the effectiveness of our proposed refinements on the vanilla model. In summary, our contributions are as follows:

\begin{itemize}
    \item We propose a framework that can reconstruct NeRF from inputs with Rolling Shutter (RS) effects by modeling the camera trajectory and aggregating RGB rendering results of different poses during the exposure time.
    \item By analyzing the vanilla framework in detail, we proposed a series of improvements to it, including trajectory smoothness regularization and the multi-sampling algorithm. 
    \item We contribute synthetic and real datasets to evaluate our framework and future works.
\end{itemize}

\section{Related Work}
\label{sec:related}

\begin{figure*}[t]
\centering
\includegraphics[width=\linewidth]{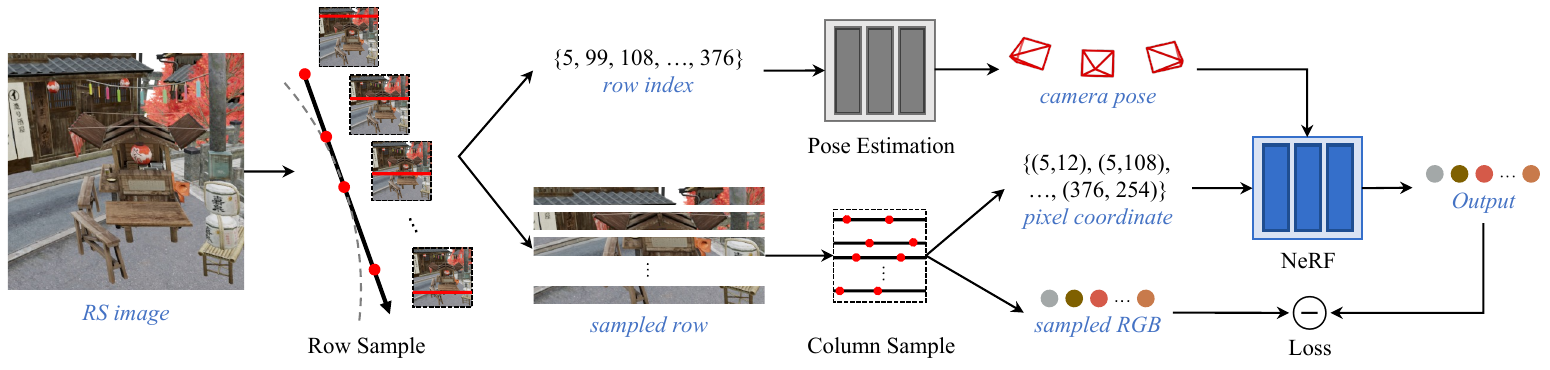}
\setlength{\abovecaptionskip}{0mm}
\caption{
\textbf{Pipeline of vanilla RS-NeRF.} Given a series of images with RS distortion, RS-NeRF learns the underlying normal 3D representations by jointly estimating pose and RGB values for each row. For each RS image, we first sample the rows and obtain their camera poses via the pose estimation network. We then sample points for each row and feed them with the estimated camera pose to the NeRF network and optimize the photo-metric loss with ground truths.
}
\label{fig:pipeline}
\end{figure*}

\noindent \textbf{2D Rolling Shutter Correction.} 
Rolling Shutter Correction (RSC) aims to restore normal images given the input with RS distortions. In recent years, several appealing image-based RSC methods~\cite{rengarajan2017unrolling,vasu2018occlusion,zhuang2019learning,liu2020deep,zhong2021towards,zhou2022evunroll,zhong2022bringing} have been developed. These methods take either single RS image or continuous RS frames as input and predict their global shutter (GS) counterparts. 
However, single-image RSC can be seriously ill-posed and struggle to correctly estimate the GS counterpart in complex scenarios. To reduce the ill-poseness of single-image RSC, more attention has been paid to multi-image RSC~\cite{liu2020deep,fan2021sunet,zhong2021towards,cao2022learning,fan2022context,fan2023joint}. 
However, these multi-image RS correction methods, which are built on image space operations, fail to exploit the 3D geometry of the scene. As a result, these methods produce multi-view inconsistency results which largely impact the reconstruction results of NeRF. 
In contrast, our proposed method can jointly reconstruct 3D geometry and correct RS images by effectively aggregating multi-view information of the scene.

\noindent \textbf{Neural Radiance Fields (NeRF).} NeRF~\cite{mildenhall2021nerf} is an end-to-end model that synthesizes novel views of complex 3D scenes by optimizing an implicit continuous volumetric scene function encoded by an MLP. Due to its simplicity in training and high-quality rendering results, NeRF-based methods~\cite{liu2020neural,garbin2021fastnerf,reiser2021kilonerf,xu2022point,chen2022aug,verbin2022refnerf,barron2021mip,barron2022mip} have attracted intensive attention from various computer vision fields, leading to significant improvements in terms of both accuracy and efficiency. 
NeRF-based models have also inspired many subsequent works that extend their continuous neural representation to different setups~\cite{guo2022nerfren,dave2022pandora,klenk2023nerf,attal2021torf,ma2022deblur,wang2023bad}. 
Deblur-NeRF~\cite{ma2022deblur} and BAD-NeRF~\cite{wang2023bad} reconstruct sharp images given the blurry input from different view directions. 
One concurrent work~\cite{li2023usb} appears in ICLR'24, which formulates the RS process with B-cubic models. This work combines the basic modeling of camera motion with the NeRF network to consider the RS effect, without further technical improvements. In this paper, we first bake the RS model into NeRF formulation, then analyze several inherit limitations of the vanilla model, and propose a series of algorithms to further improve its performance.

\section{RS-NeRF}
\label{sec:method}

In this section, we present the specifics of RS-NeRF. Initially, we explain the basic formulation of RS-NeRF in Sec.~\ref{sec:rs_model}. We also note that the vanilla model possesses specific shortcomings that hinder its effectiveness. To address these, we suggest several improvements to enhance its capabilities, which are detailed in Sec.~\ref{sec:smooth} and Sec.~\ref{sec:multi}.

\subsection{Vanilla RS-NeRF Model}
\label{sec:rs_model}

\noindent \textbf{Rolling shutter camera model.} Unlike Global Shutter (GS) cameras, the Rolling Shutter (RS) camera captures each row at distinct timestamps. Without loss of generality, we consider the readout direction of the RS camera to be from top to bottom, assuming an infinitesimal exposure time for each row, following existing RS modeling techniques~\cite{liu2020deep,fan2022context,fan2023joint}. This mechanism can be described through a mathematical model as follows:
\begin{align}
    [\mathbf{I}^R(\mathbf{x})]_i = [\mathbf{I}^G_i(\mathbf{x})]_i,
\end{align}
where $[\mathbf{I}(\mathbf{x})]_i$ is an operator to extract the $i$-th row from an image $\mathbf{I}(\mathbf{x})$, $\mathbf{I}^R(\mathbf{x}) \in \mathbb{R}^{H \times W \times 3}$ is the rolling shutter image, and $\mathbf{I}_i^G(\mathbf{x}) \in \mathbb{R}^{H \times W \times 3}$ is the virtual global shutter image captured at the same pose as the $i$-th row of $\mathbf{I}^R(\mathbf{x})$. As a result, for one RS image $\mathbf{I}^R(\mathbf{x})$, we denote the pose of the $i$-th row as $\mathbf{T}_i$.

\noindent \textbf{Camera motion trajectory modeling.} In our rolling shutter camera model, it is necessary to establish the poses for each row in each image. Given the typically short exposure time of an RS image, we use a linear model to approximate the motion of the camera. Specifically, we start by defining two camera poses: one for the first row, denoted as $\mathbf{T}_0$, and one for the last row, $\mathbf{T}_{H-1}$. For the rows in between, we linearly interpolate their poses within the Lie algebra of $\mathbf{S}\mathbf{E}(3)$, as informed by prior research~\cite{lin2021barf,wang2023bad,liu2021mba}:
\begin{align}
    \mathbf{T}_i = \mathbf{T}_0 \cdot (1 - \rho) + \mathbf{T}_{H-1} \cdot \rho,
\end{align}
where the interpolation coefficient $\rho = \frac{i}{H-1}$ is based on the row index $i$. Thus, for each RS image, RS-NeRF receives one pose initialization, then optimizes the poses at both ends and then determines the poses for each intermediate row through linear interpolation. This interpolation method for calculating each row's pose is fully differentiable, enabling the optimization of pose parameters using a series of rolling shutter images.

\noindent \textbf{Optimizing NeRF from RS observation.} For a set of $K$ rolling shutter images, we select $M$ rows per image and $N$ points per row, resulting in $K \times M \times N$ distinct camera rays for each iteration. These rays are used to concurrently optimize the NeRF network and the pose parameters. Specifically, for each image, we initially use the row indices to access the pose estimation network. This allows us to determine the camera ray $\hat{\mathbf{r}}(\theta_p; u,v))$ for each sampled pixel point $(u, v)$. Following this, these rays are employed to train the NeRF network by computing the ensuing photometric loss:
\begin{align}
    \mathcal{L} = ||\mathbf{C}(\mathbf{r}(u,v)) - \hat{\mathbf{C}}(\theta_n; \hat{\mathbf{r}}(\theta_p; (u,v))||_2,
\end{align}
where $\theta_n$ represents the trainable parameters of the NeRF network, and $\theta_p$ denotes the learnable parameters for camera poses. $\hat{\mathbf{C}}(\theta_n; \hat{\mathbf{r}}(\theta_p; (u,v))$ refers to the predicted RGB value for the pixel at coordinates $(u,v)$, while $\mathbf{C}(\mathbf{r}(u,v))$ denotes the corresponding RGB ground truth value. The workflow of our vanilla RS-NeRF model is depicted in Fig.~\ref{fig:pipeline}.

{
\sidecaptionvpos{figure}{t}
\begin{SCfigure}[][t]
\includegraphics[width=0.5\textwidth]{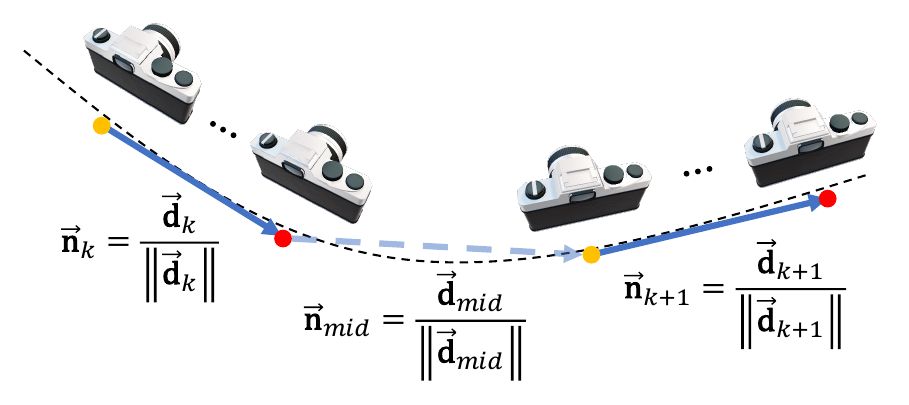}
\setlength{\abovecaptionskip}{0mm}
\caption{
\textbf{Smoothness regularization on camera trajectory.} For each pair of adjacent trajectory vectors $\overrightarrow{\mathbf{d}}_{k}$ and $\overrightarrow{\mathbf{d}}_{k+1}$, we compute the mid-point trajectory vector $\overrightarrow{\mathbf{d}}_{mid}$ and their unit vectors $\overrightarrow{\mathbf{n}}_{k}$, $\overrightarrow{\mathbf{n}}_{k+1}$, and $\overrightarrow{\mathbf{n}}_{mid}$. We then apply the $L_2$ regularization between $\overrightarrow{\mathbf{n}}_{mid}$ and $\operatorname{mean}(\overrightarrow{\mathbf{n}}_{k}, \overrightarrow{\mathbf{n}}_{k+1})$.
}
\label{fig:smooth}
\end{SCfigure}
}


\subsection{Trajectory Smoothness Regularization.} 
\label{sec:smooth}

Given the previous description of our model, it is evident that accurately estimating the camera pose for each row in RS frames is crucial for effectively reconstructing the normal NeRF representation. We observe that the basic RS-NeRF struggles with precise camera trajectory estimation, often resulting in high-frequency predictions, as further analyzed in our experiments. This issue arises because the network focuses only on the photometric loss for individual images, neglecting the consistent movement characteristics of the camera across different RS frames. To address this, we introduce a camera trajectory smoothness regularization, based on the expectation of a smooth and continuous camera trajectory in 3D space. Specifically, for every two adjacent RS frames $\mathbf{I}^R_k$ and $\mathbf{I}^R_{k+1}$, we aim to minimize the following loss during the optimization process:
\begin{align}
    \mathcal{L}_{s}=||\overrightarrow{\mathbf{n}}_{mid} - \operatorname{mean}(\overrightarrow{\mathbf{n}}_k + \overrightarrow{\mathbf{n}}_{k+1})||_2.
\end{align}
As illustrated in Fig.~\ref{fig:smooth}, $\overrightarrow{\mathbf{n}}_{k}$ and $\overrightarrow{\mathbf{n}}_{k+1}$ represent the unit direction vectors of the estimated camera trajectories for RS frames $\mathbf{I}^R_k$ and $\mathbf{I}^R_{k+1}$, respectively. Meanwhile, $\overrightarrow{\mathbf{n}}_{mid}$ denotes the unit direction vector for the camera trajectory during the exposure interval between these two RS frames.

\begin{table*}[t] 
\centering
\begin{minipage}[b]{0.42\textwidth}
\centering
\resizebox{\linewidth}{!}{%
\begin{tabular}{c|c|c|c}
\toprule
\multirow{2}{*}{$N_{pose}$} & \multicolumn{3}{c}{\textsc{On Traj. / Out Traj.}} \\
 & \small{PSNR $\uparrow$} & \small{SSIM $\uparrow$} & \small{LPIPS $\downarrow$} \\
\midrule
$100$ & 24.36/25.16 & .7321/.7727 & .0703/.0578 \\
$400$ & 24.06/24.66 & .7245/.7536 & .0697/.0579 \\
\bottomrule
\end{tabular}%
}
\setlength{\abovecaptionskip}{2mm}
\caption{\textbf{Quantitative results for different $N_{pose}$.} $N_{pose}=100$ yields better results than $N_{pose}=400$.}
\label{tab:pose_number}
\end{minipage}
\hfill 
\begin{minipage}[b]{0.57\textwidth}
\centering
\includegraphics[width=\textwidth]{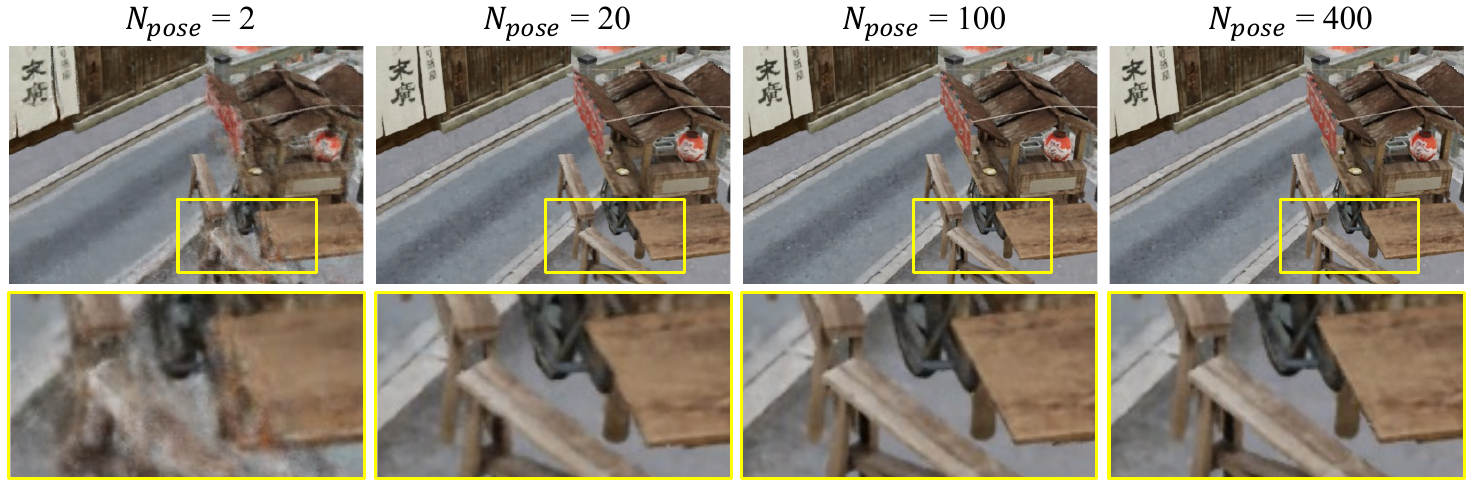}
\setlength{\abovecaptionskip}{-2mm}
\captionof{figure}{\textbf{Qualitative comparisons for different $N_{pose}$.} The artifact decreases as $N_{pose}$ grows.}
\label{fig:toy}
\end{minipage}
\end{table*}


\subsection{Multi-Sampling Algorithm}
\label{sec:multi}

As discussed in Sec.~\ref{sec:rs_model}, each row in an RS frame is assigned a unique camera pose. For example, in an RS frame with 400 rows, there are 400 distinct camera poses. This rigorous approach aligns with the real-world, physics-based photography process of RS cameras but also presents certain challenges. For example, under this framework, the data points for each pose are confined to a single row. This limitation can significantly affect our model's performance due to the sparse training data available for each pose.

\begin{figure*}[t]
\centering
\includegraphics[width=0.8\linewidth]{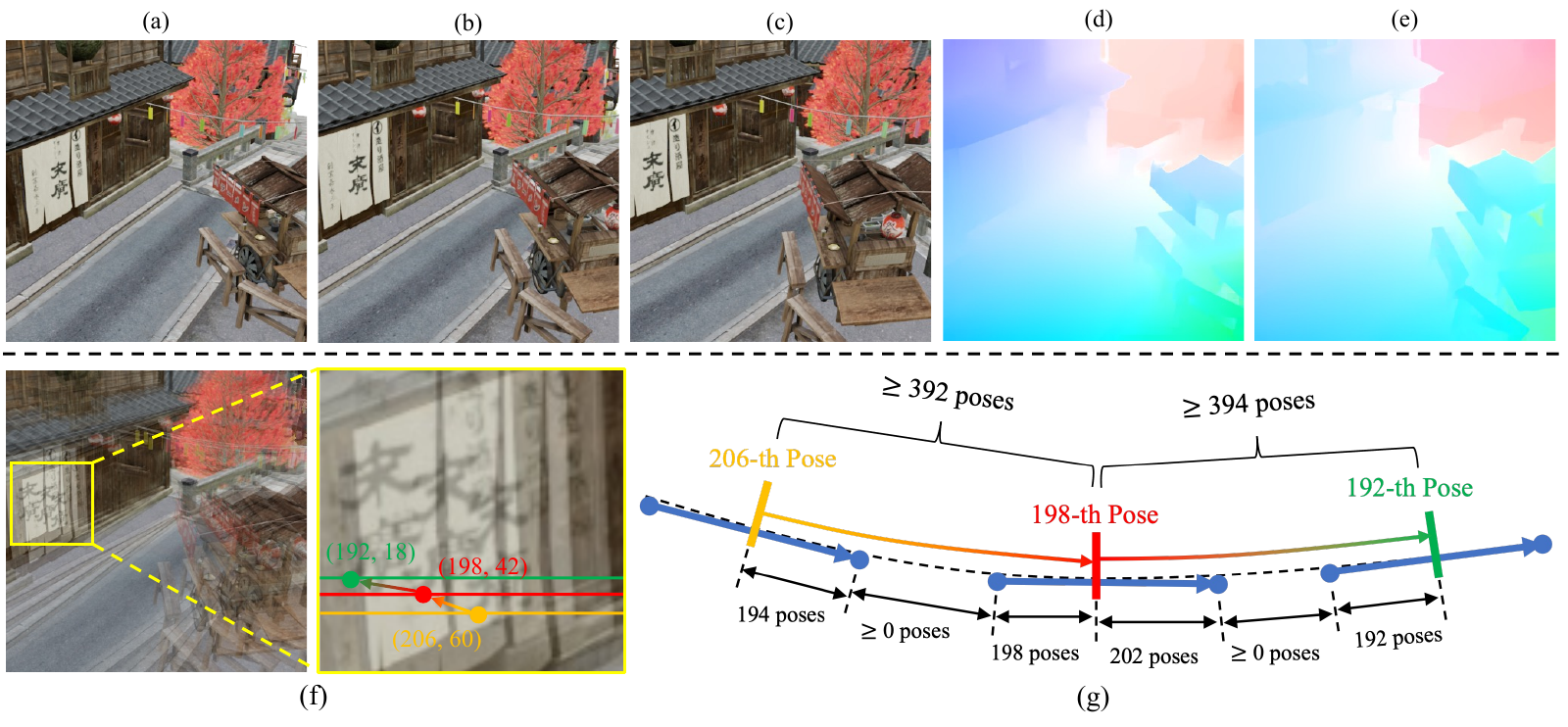}
\caption{
\textbf{Motivation for the multi-sampling algorithm.} \textbf{Up:} Three consecutive RS frames, labeled as (a), (b), and (c), along with the optical flow between each pair of frames, shown in (d) and (e).
\textbf{Down:} For a single point in 3D space, we observe its 2D projections in the first (marked in yellow) and second (marked in red) RS frames. These projections exhibit a row displacement of 8 and a column displacement of 18, as depicted in (f). However, during this interval, the camera undergoes a shift of at least 392 poses, as indicated in (g).
}
\label{fig:growing}
\end{figure*}


A straightforward solution to this problem is to reduce the number of poses per image. For example, assigning 200 poses to each RS frame means that all two adjacent rows would share the same pose. This approach can lead to blurriness since these rows are captured at different poses, but it also doubles the amount of data available for training each pose. To further explore this trade-off, we conducted a series of experiments. Fig.~\ref{fig:toy} shows qualitative results for different intermediate pose numbers $N_{pose}$, indicating that the artifacts decrease as $N_{pose}$ increases. Interestingly, we observed that using 100 poses per frame often yields results comparable to or even better than using 400 poses. The quantitative results presented in Tab.~\ref{tab:pose_number} also support the superiority of having 100 poses per frame.

This observation seems counter-intuitive at first, as one might assume that having fewer than 400 poses per frame would lead to poorer results due to artifacts like blurriness. However, it can be rationally explained. As demonstrated in Fig.~\ref{fig:growing}, for a given point in 3D space, the 2D projections on the first and second RS frames show relatively small pixel displacements (8 in row and 12 in column). Meanwhile, the camera shifts through at least 392 poses in 3D space. This is based on the assumption that there is no overlap in the trajectories of two adjacent frames (indicated as '$\geq$ 0 poses' in the figure), which is naturally satisfied given the sparse observation setting and short exposure times typically involved.

Following our prior analysis, we introduced a multi-sampling algorithm aimed at explicitly increasing the sample points for each intermediate camera pose. The fundamental concept involves computing the ratio of camera pose displacement to pixel displacement (termed the PP-ratio) between two adjacent frames for the same 3D point. For example, a pixel displacement of 18 combined with a camera pose shift of 392 results in a PP-ratio of 21. This implies that at least 21 consecutive camera poses share the same pixel value, which means that all these 21 camera poses can utilize this RGB value for training purposes.

For a pixel $\mathbf{x}=(u,v)$ located in the $u$-th column and $v$-th row of the $k$-th RS frame $\mathbf{I}^R_{k}(\mathbf{x})$, we determine both the forward PP-ratio $\rho_{k\rightarrow{k+1}}(\mathbf{x})$ and the backward PP-ratio $\rho_{k\rightarrow{k-1}}(\mathbf{x})$. This is because the point represented by $\mathbf{x}$ can transition from $\mathbf{I}^R_{k}(\mathbf{x})$ to $\mathbf{I}^R_{k+1}(\mathbf{x})$ or $\mathbf{I}^R_{k-1}(\mathbf{x})$. Consequently, all camera poses within the range of $[v-\rho_{k\rightarrow{k-1}}(\mathbf{x}), v+\rho_{k\rightarrow{k+1}}(\mathbf{x})]$ are capable of sampling the RGB value of $\mathbf{x}$ for training purposes.

To compute the forward PP-ratio $\rho_{k\rightarrow{k+1}}(\mathbf{x})$ for a pixel $\mathbf{x}$ in the $k$-th RS frame $\mathbf{I}^R_{k}(\mathbf{x})$, the first step is to estimate the forward optical flow $F_{k\rightarrow{k+1}}(\mathbf{x})$, utilizing RAFT~\cite{teed2020raft}. Consequently, the row displacement $\Delta^{row}_{k\rightarrow{k+1}}(\mathbf{x})$ and column displacement $\Delta^{col}_{k\rightarrow{k+1}}(\mathbf{x})$ of $\mathbf{x}$ from $\mathbf{I}^R_{k}(\mathbf{x})$ to $\mathbf{I}^R_{k+1}(\mathbf{x})$ would be the second and the first element of $F_{k\rightarrow{k+1}}(\mathbf{x})$:
\begin{align}
    F_{k\rightarrow{k+1}}(\mathbf{x}) = \operatorname{RAFT}\left(\mathbf{I}^R_{k}(\mathbf{x}), \mathbf{I}^R_{k+1}(\mathbf{x})\right),  \\
    \Delta^{col}_{k\rightarrow{k+1}}(\mathbf{x}), \Delta^{row}_{k\rightarrow{k+1}}(\mathbf{x}) = F_{k\rightarrow{k+1}}(\mathbf{x}),
\end{align}
and the calculation are similar for the backward situation.

We then calculate the forward and backward camera pose shift as:
\begin{align}
    \Delta^{pose}_{k\rightarrow{k+1}}(\mathbf{x}) &= v + \Delta^{row}_{k\rightarrow{k+1}}(\mathbf{x}) + (H - v), \\
    &= H + \Delta^{row}_{k\rightarrow{k+1}}(\mathbf{x}), \\
    \Delta^{pose}_{k\rightarrow{k-1}}(\mathbf{x}) &= \left(H - (v + \Delta^{row}_{k\rightarrow{k-1}}(\mathbf{x}))\right) + v, \\
    &= H - \Delta^{row}_{k\rightarrow{k-1}}(\mathbf{x}).
\end{align}

Finally, we calculate forward and backward PP-ratio as:
\begin{align}
    \rho_{k\rightarrow{k+1}}(\mathbf{x}) = \frac{\Delta^{pose}_{k\rightarrow{k+1}}(\mathbf{x})}{\max(\Delta^{col}_{k\rightarrow{k+1}}(\mathbf{x}), \Delta^{row}_{k\rightarrow{k+1}}(\mathbf{x}))}, \\
    \rho_{k\rightarrow{k-1}}(\mathbf{x}) = \frac{\Delta^{pose}_{k\rightarrow{k-1}}(\mathbf{x})}{\max(\Delta^{col}_{k\rightarrow{k-1}}(\mathbf{x}), \Delta^{row}_{k\rightarrow{k-1}}(\mathbf{x}))}.
\end{align}
Therefore, when dealing with a sampled pixel $RGB(\mathbf{x})$ in the RS frame $\mathbf{I}^R_{k}(\mathbf{x})$, where $\mathbf{x}=(u,v)$, rather than directly using the $v$-th pose $\mathbf{T}_v$ of $\mathbf{I}^R_{k}(\mathbf{x})$ for training, we randomly sample $\mathbf{T}_{\hat{v}}$ where
\begin{align}
    \hat{v} \in [v-\rho_{k\rightarrow{k-1}}(\mathbf{x}), v+\rho_{k\rightarrow{k+1}}(\mathbf{x})].
\end{align}

{
\sidecaptionvpos{figure}{t}
\begin{SCfigure}[][t]
\includegraphics[width=0.6\textwidth]{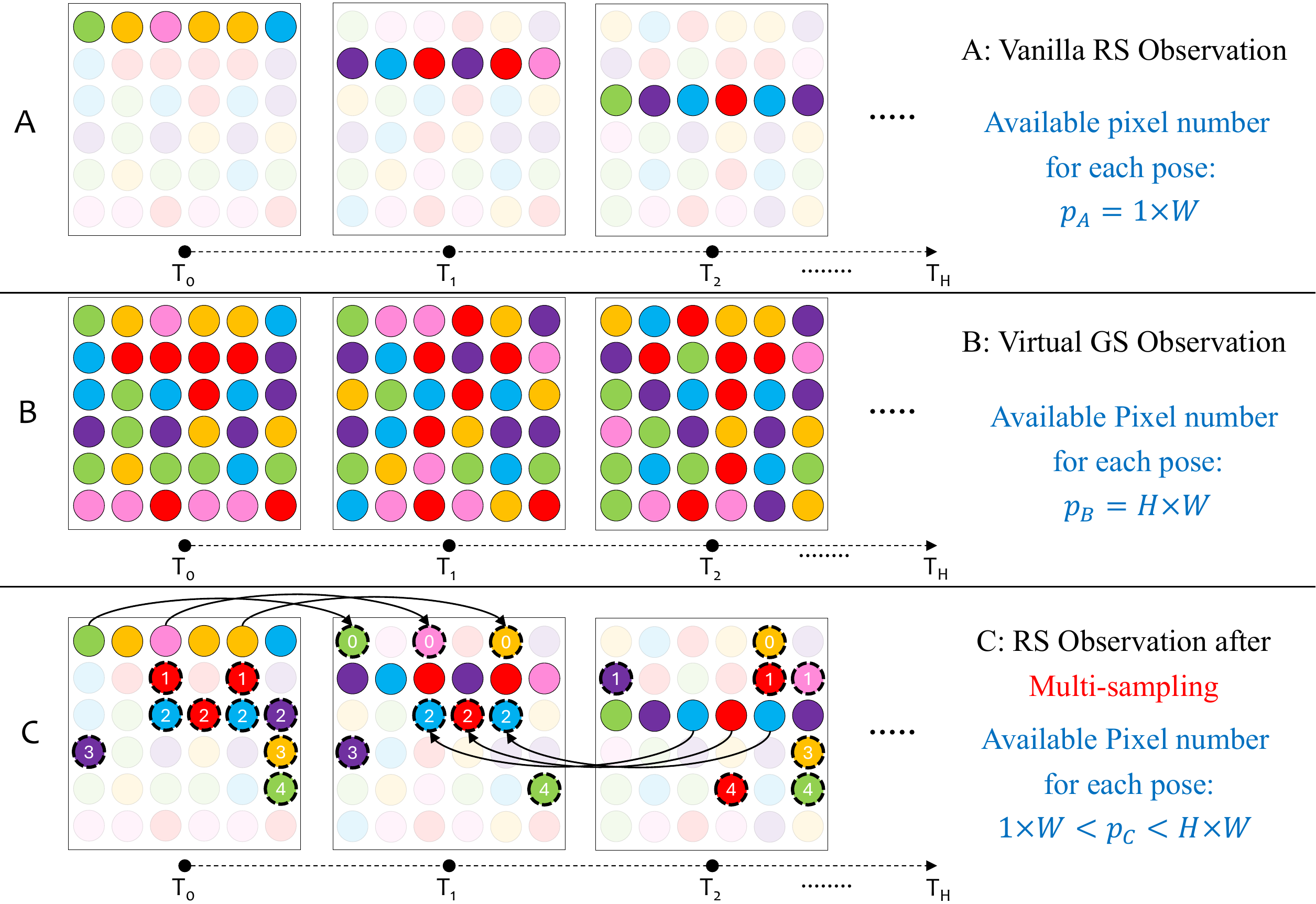}
\caption{
\textbf{Comparing the accessible training points in one RS image under different strategies.} The image size is $H \times W$, providing $H$ poses. Distinct circle hues represent different RGB values. In case \textit{C}, pixels transported from other poses are with dash contours, and the number signifying that it originates from the $i$-th pose.
}
\label{fig:ms}
\end{SCfigure}
}

\noindent \textbf{Why does Multi-Sampling work?} 
Let us consider a basic task: optimizing NeRF with $H$ known poses and images (non-RS). Case \textit{A}: Only one row is available per pose. Case \textit{B}: all rows are accessible. Clearly, case \textit{B} would yield a better NeRF due to more data per pose. Moving to our RS setting as depicted in Fig.\ref{fig:ms}, case \textit{A} is the vanilla RS observation for one RS frame ($H$ rows lead to $H$ poses), and case \textit{B} is the \textit{virtual} GS observation, where all rows are available for each pose. Thus, case \textit{B} is the upper bound for case \textit{A}. Essentially, MS aims to \textit{transport all possible pixel values from nearby poses to fill the missing observation that can be potentially recovered} (case \textit{C} in Fig.\ref{fig:ms}), \ie, \textit{partially restore the virtual GS observation from the vanilla RS observation, by allocating additional poses for each pixel}, which ensures better results than vanilla RS. Also, interpolating two poses does not harm MS. We optimize two poses for each frame, and each in-between pose becomes a fixed blend ratio of two poses. The best case is virtual GS (Fig.\ref{fig:ms}(B)), where every pixel coordinate can be sampled in all ratios ($\frac{0}{H}, \frac{1}{H-1}, ..., \frac{H}{0}$), resulting in a total of \textit{H}$\times$\textit{W}$\times$\textit{H} data points to optimize the two poses. In vanilla RS (Fig.\ref{fig:ms}(A)), each coordinate only contributes to one ratio determined by row indices, producing only \textit{1}$\times$\textit{W}$\times$\textit{H} data points. With MS (Fig.\ref{fig:ms}(C)), some pixels are replicated to other poses, allowing multiple ratios for them. Thus, the number of data points is greater than \textit{1}$\times$\textit{W}$\times$\textit{H} but less than \textit{H}$\times$\textit{W}$\times$\textit{H}. This ensures better pose estimation and view synthesis than vanilla RS.

\begin{table*}[t]
\centering
\setlength{\tabcolsep}{0.8mm}
\resizebox{\textwidth}{!}{%
\begin{tabular}{c|cc|ccccccccccccccc|ccc}
\toprule
& \multirow{2}{*}{$\mathcal{L}_{s}$} & \multirow{2}{*}{MS} & \multicolumn{3}{c}{\normalsize{\textsc{Torii}}} & \multicolumn{3}{c}{\normalsize{\textsc{Wine}}} & \multicolumn{3}{c}{\normalsize{\textsc{Pool}}} & \multicolumn{3}{c}{\normalsize{\textsc{Factory}}} & \multicolumn{3}{c}{\normalsize{\textsc{Tanabata}}} & \multicolumn{3}{|c}{\normalsize{\textsc{Average}}} \\
& &  & \small{PSNR $\uparrow$} & \small{SSIM $\uparrow$} & \small{LPIPS $\downarrow$} & \small{PSNR $\uparrow$} & \small{SSIM $\uparrow$} & \small{LPIPS $\downarrow$} & \small{PSNR $\uparrow$} & \small{SSIM $\uparrow$} & \small{LPIPS $\downarrow$} & \small{PSNR $\uparrow$} & \small{SSIM $\uparrow$} & \small{LPIPS $\downarrow$} & \small{PSNR $\uparrow$} & \small{SSIM $\uparrow$} & \small{LPIPS $\downarrow$} & \small{PSNR $\uparrow$} & \small{SSIM $\uparrow$} & \small{LPIPS $\downarrow$} \\
\midrule
\multirow{4}{*}{\rotatebox[origin=c]{90}{\textsc{On Traj.}}} & &  & 23.45 & .7535 & .0448 & 23.53 & .7019 & \secondscore{.0681} & 22.99 & .6111 & .0784 & 19.53 & .5018 & .1072 & 17.86 & .4921 & .1030 & 21.47 & .6121 & .0803 \\
 & \Checkmark &  & 22.25 & .6729 & .0564 & \secondscore{25.60} & \secondscore{.7749} & .0712 & \secondscore{27.22} & \secondscore{.7705} & .0582 & 21.81 & .6128 & .0949 & \secondscore{23.42} & \secondscore{.7916} & \secondscore{.0679} & 24.06 & .7245 & .0697 \\
 & & \Checkmark & \secondscore{26.65} & \secondscore{.8572} & \bestscore{.0342} & 25.55 & .7637 & .0682 & 25.86 & .7303 & \secondscore{.0555} & \bestscore{25.45} & \bestscore{.7754} & \bestscore{.0708} & 20.94 & .6802 & .0832 & \secondscore{24.89} & \secondscore{.7614} & \bestscore{.0624} \\
 & \Checkmark & \Checkmark & \bestscore{30.97} & \bestscore{.9058} & \secondscore{.0395} & \bestscore{28.36} & \bestscore{.8482} & \bestscore{.0675} & \bestscore{28.50} & \bestscore{.8137} & \bestscore{.0562} & \secondscore{25.43} & \secondscore{.7673} & \secondscore{.0867} & \bestscore{26.40} & \bestscore{.8614} & \bestscore{.0651} & \bestscore{27.93} & \bestscore{.8393} & \secondscore{.0630} \\
\midrule
\multirow{4}{*}{\rotatebox[origin=c]{90}{\textsc{Out Traj.}}} &  &  & 23.26 & .7561 & .0428 & 24.05 & .7169 & .0601 & 23.27 & .6218 & .0738 & 19.98 & .5426 & .0892 & 17.85 & .5065 & .0942 & 21.68 & .6288 & .0720 \\
 & \Checkmark &  & 22.39 & .6875 & .0501 & \secondscore{26.62} & \secondscore{.8113} & \secondscore{.0579} & \secondscore{27.29} & \secondscore{.7783} & \secondscore{.0489} & 22.72 & .6624 & .0815 & \secondscore{24.27} & \secondscore{.8286} & \secondscore{.0514} & \secondscore{24.66} & .7536 & \secondscore{.0579} \\
 &  & \Checkmark & \secondscore{26.58} & \secondscore{.8611} & \secondscore{.0308} & 25.96 & .7750 & .0612 & 25.49 & .7080 & .0549 & \secondscore{24.38} & \secondscore{.7446} & \secondscore{.0742} & 20.73 & .6850 & .0751 & 24.63 & \secondscore{.7547} & .0592 \\
 & \Checkmark & \Checkmark & \bestscore{33.07} & \bestscore{.9499} & \bestscore{.0243} & \bestscore{29.27} & \bestscore{.8703} & \bestscore{.0560} & \bestscore{28.77} & \bestscore{.8313} & \bestscore{.0458} & \bestscore{25.11} & \bestscore{.7738} & \bestscore{.0807} & \bestscore{27.02} & \bestscore{.8881} & \bestscore{.0498} & \bestscore{28.65} & \bestscore{.8627} & \bestscore{.0513} \\
\bottomrule
\end{tabular}%
}
\setlength{\abovecaptionskip}{2mm}
\caption{\textbf{Quantitative results on synthetic scenes for ablation study.} $\mathcal{L}_{s}$ denotes the camera trajectory smoothness regularization, and `MS' denotes the multi-sampling algorithm. We color code each result as \bestscore{best} and \secondscore{second best}.}
\label{tab:ablation}
\end{table*}

\section{Experiments}
\label{sec:experiments}

\subsection{Settings}

\noindent \textbf{Datasets.} Evaluating RS-aware view synthesis technologies requires synthetic datasets that have multi-view images with RS distortions and the corresponding GS ground truth. We synthesize 5 scenes via blender to evaluate our method. For each scene, we rendered images from various camera poses while moving the camera gradually around the scene in a forward-facing manner. We then combined rows from different poses to produce the final RS images. Each scene comprises 34 images with a resolution of $400 \times 400$. We employed the pose of the middle row of each RS image as the initial camera pose for our model. We also analyze our model's robustness to pose initialization in the experiment section. For evaluation, we used 34 on-trajectory views and 16 out-trajectory views to test RS correction and novel view synthesis results, respectively. To test our model in real-world scenarios, we also captured real scenes using the RS camera EO-1312C. These RS images were captured by manually moving the camera during exposure, in a forward-facing manner, with a resolution of $400 \times 500$. We utilize COLMAP~\cite{schonberger2016structure,schonberger2016pixelwise} for the initial camera pose in our model.

\noindent \textbf{Evaluation Metrics.} The quality of the rendered image is evaluated with the commonly used metrics including PSNR, SSIM and LPIPS~\cite{zhang2018unreasonable}. To evaluate the accuracy of the estimated camera trajectory, we calculate the Mean Square Trajectory Error (MSTE) with the ground-truth camera trajectory.

\subsection{Implementation Details}

In each training iteration, we randomly choose 64 rows from each of the images and 2 pixels from each of these rows. We employ the Adam optimizer~\cite{kingma2014adam} with $\beta_1=0.9$ and $\beta_2=0.999$ to optimize our model. The learning rate is initially set at $5 \times 10^{-4}$ and undergoes exponential decay throughout the optimization process. To avoid potential errors from inaccurate optical flow estimation, we disable multi-sampling and use the vanilla sampling strategy in the later part of the training process. This makes the model use much fewer but absolutely accurate data points in the final training phase. Each scene is trained for 200,000 iterations. All this training is conducted on a single NVIDIA 4090 GPU.

\begin{figure*}[t]
\centering
\includegraphics[width=\linewidth]{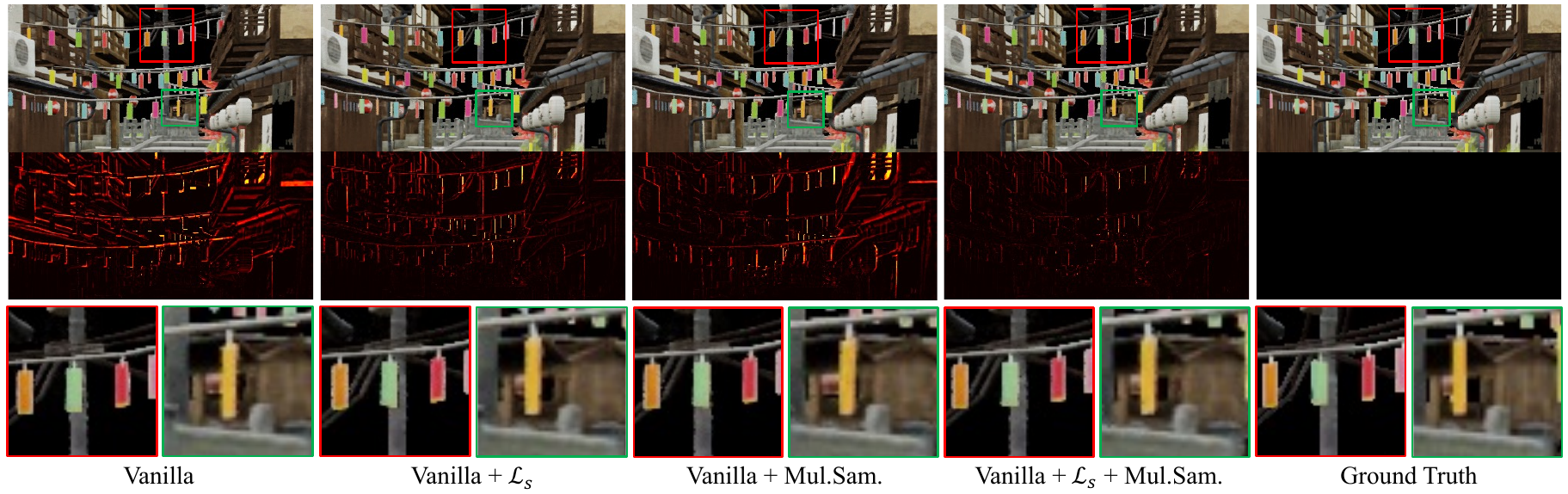}
\caption{\textbf{Qualitative results on synthetic scenes for ablation study.} The corresponding error maps and zoomed regions are visualized in the bottom. Our full model has the smaller error to the GT, with less spatial misalignment and twist caused by the RS distortion.}
\label{fig:ablation}
\end{figure*}

{
\sidecaptionvpos{table}{t}
\begin{SCtable}[][t]
\resizebox{0.55\linewidth}{!}{%
\begin{tabular}{c|c|c|c}
\midrule
Vanilla & Vanilla+$\mathcal{L}_s$ & Vanilla+MS & Vanilla+$\mathcal{L}_s$+MS \\
\hline
0.0509 & 0.0358 & 0.0462 & \textbf{0.0319} \\
\midrule
\end{tabular}%
}
\caption{
\textbf{Average MSTE (the lower the better) results} for ablation study.
}
\label{tab:ate_ablation}
\end{SCtable}
}

\subsection{Ablation Study}
To comprehensively evaluate the effectiveness of each component in our RS-NeRF model, we carried out extensive ablation studies. The quantitative and qualitative results are presented in Tab.~\ref{tab:ablation}, Fig.~\ref{fig:ablation} and Tab.~\ref{tab:ate_ablation}, respectively.

\noindent \textbf{Trajectory Smoothness Regularization.} The accurate estimation of camera trajectory is crucial for effectively restoring normal images from RS inputs, given the inherent formation process of RS effects. As evidenced in Tab.~\ref{tab:ablation}, RS-NeRF enhanced with trajectory smoothness regularization ($\mathcal{L}_{s}$) shows notable improvements over the vanilla model. Specifically, it achieves a PSNR increase of $2.59$ for on-trajectory views and $2.98$ for out-trajectory views. Additionally, removing $\mathcal{L}_{s}$ from our full model leads to a PSNR decrease of $3.04$ for on-trajectory views and $4.02$ for out-trajectory views. The qualitative results further underscore the effectiveness of $\mathcal{L}_{s}$. As illustrated in Fig.~\ref{fig:ablation}, the vanilla RS-NeRF, when complemented with $\mathcal{L}_{s}$, exhibits reduced spatial misalignment and a smaller synthesis error to the ground truth. Furthermore, Tab.~\ref{tab:ate_ablation} reveals that the vanilla RS-NeRF tends to provide an inaccurate trajectory estimation, primarily focusing on the photo-metric loss for each image and overlooking the pattern of camera movement across different RS frames. In contrast, the `Vanilla + $\mathcal{L}_{s}$' model, with the addition of smoothness regularization, yields a more accurate estimation of the camera trajectory, resulting in improved synthesis outcomes.

\noindent \textbf{Multi-Sampling Algorithm.} The Multi-sampling algorithm in RS-NeRF significantly enhances the training data available for each intermediate pose by calculating the PP-ratio for each pixel, marking a notable advancement over the vanilla model. According to the results in Tab.~\ref{tab:ablation}, RS-NeRF equipped with the multi-sampling algorithm shows substantial improvements: a PSNR increase of $3.42$ for on-trajectory views and $2.95$ for out-trajectory views. Moreover, excluding the multi-sampling algorithm from our full model leads to a PSNR reduction of $3.87$ for on-trajectory views and $3.99$ for out-trajectory views. The qualitative results also affirm the effectiveness of it. As depicted in Fig.~\ref{fig:ablation}, our model, with the multi-sampling algorithm, displays fewer artifacts and a smaller synthesis error to the ground truth. Interestingly, the multi-sampling algorithm also seems to enhance trajectory estimation. As shown in Tab.~\ref{tab:ate_ablation}, RS-NeRF with only the multi-sampling algorithm (Vanilla + MS) provides a more accurate trajectory estimation than the vanilla model, which likely contributes to its significantly improved synthesis results.

\begin{table*}[t]
\centering
\setlength{\tabcolsep}{0.8mm}
\resizebox{\textwidth}{!}{%
\begin{tabular}{c|l|ccccccccccccccc|ccc}
\toprule
& \multirow{2}{*}{Methods} & \multicolumn{3}{c}{\normalsize{\textsc{Torii}}} & \multicolumn{3}{c}{\normalsize{\textsc{Wine}}} & \multicolumn{3}{c}{\normalsize{\textsc{Pool}}} & \multicolumn{3}{c}{\normalsize{\textsc{Factory}}} & \multicolumn{3}{c}{\normalsize{\textsc{Tanabata}}} & \multicolumn{3}{|c}{\normalsize{\textsc{Average}}} \\
& & \small{PSNR $\uparrow$} & \small{SSIM $\uparrow$} & \small{LPIPS $\downarrow$} & \small{PSNR $\uparrow$} & \small{SSIM $\uparrow$} & \small{LPIPS $\downarrow$} & \small{PSNR $\uparrow$} & \small{SSIM $\uparrow$} & \small{LPIPS $\downarrow$} & \small{PSNR $\uparrow$} & \small{SSIM $\uparrow$} & \small{LPIPS $\downarrow$} & \small{PSNR $\uparrow$} & \small{SSIM $\uparrow$} & \small{LPIPS $\downarrow$} & \small{PSNR $\uparrow$} & \small{SSIM $\uparrow$} & \small{LPIPS $\downarrow$} \\
\midrule
\multirow{4}{*}{\rotatebox[origin=c]{90}{\textsc{On Traj.}}}
& \textit{DU} & \secondscore{21.23} & \secondscore{.6409} & \secondscore{.0845} & \secondscore{19.87} & \secondscore{.4520} & \secondscore{.1329} & \secondscore{23.08} & \secondscore{.5394} & \secondscore{.1189} & \secondscore{19.95} & \secondscore{.5128} & \secondscore{.1325} & \secondscore{18.17} & .5262 & \secondscore{.1213} & \secondscore{20.46} & \secondscore{.5343} & \secondscore{.1180} \\
& \textit{CVR} & 19.94 & .6148 & .1454 & 18.86 & .4090 & .2052 & 21.51 & .5173 & .1885 & 18.59 & .4851 & .1883 & 17.45 & .5277 & .1848 & 19.27 & .5108 & .1824 \\
& \textit{JAM} & 20.66 & .6311 & .0872 & 19.50 & .4214 & .1595 & 22.43 & .5180 & .1263 & 19.34 & .4907 & .1510 & 17.87 & \secondscore{.5342} & .1292 & 19.96 & .5191 & .1307 \\
& RS-NeRF & \bestscore{30.97} & \bestscore{.9058} & \bestscore{.0395} & \bestscore{28.36} & \bestscore{.8482} & \bestscore{.0675} & \bestscore{28.50} & \bestscore{.8137} & \bestscore{.0562} & \bestscore{25.43} & \bestscore{.7673} & \bestscore{.0867} & \bestscore{26.40} & \bestscore{.8614} & \bestscore{.0651} & \bestscore{27.93} & \bestscore{.8393} & \bestscore{.0630} \\
\midrule
\multirow{5}{*}{\rotatebox[origin=c]{90}{\textsc{Out Traj.}}} & \textit{NeRF} & 19.35 & .5976 & .2535 & 19.06 & .3845 & .3550 & 22.02 & .4740 & .3196 & 17.87 & .4075 & .3570 & 17.56 & .4958 & .3019 & 19.17 & .4719 & .3174 \\
& \textit{DU+NeRF} & 20.33 & .6585 & .2528 & 20.85 & .4923 & .3404 & 23.25 & .5403 & .2860 & 19.16 & .5037 & .3219 & 18.14 & .5226 & .3102 & 20.35 & .5435 & .3022 \\
& \textit{CVR+NeRF} & 23.67 & .7779 & .1967 & \secondscore{22.27} & \secondscore{.6114} & \secondscore{.2791} & 24.56 & .6802 & .2408 & 19.81 & .5619 & .2996 & 18.58 & .6250 & .2736 & 21.78 & .6513 & .2580 \\
& \textit{JAM+NeRF} & \secondscore{24.13} & \secondscore{.7852} & \secondscore{.1512} & 22.14 & .5801 & .2822 & \secondscore{25.51} & \secondscore{.6867} & \secondscore{.1855} & \secondscore{20.46} & \secondscore{.5834} & \secondscore{.2717} & \secondscore{18.63} & \secondscore{.6357} & \secondscore{.2388} & \secondscore{22.18} & \secondscore{.6542} & \secondscore{.2259} \\
& RS-NeRF & \bestscore{33.07} & \bestscore{.9499} & \bestscore{.0243} & \bestscore{29.27} & \bestscore{.8703} & \bestscore{.0560} & \bestscore{28.77} & \bestscore{.8313} & \bestscore{.0458} & \bestscore{25.11} & \bestscore{.7738} & \bestscore{.0807} & \bestscore{27.02} & \bestscore{.8881} & \bestscore{.0498} & \bestscore{28.65} & \bestscore{.8627} & \bestscore{.0513} \\
\bottomrule
\end{tabular}%
}
\setlength{\abovecaptionskip}{2mm}
\caption{\textbf{Quantitative comparison on synthetic scenes of different methods.} DU, CVR, and JAM represent DeepUnroll~\cite{liu2020deep}, CVR~\cite{fan2022context}, and JAMNet~\cite{fan2023joint} respectively. We color code each result as \bestscore{best} and \secondscore{second best}.}
\label{tab:comparison}
\end{table*}

\begin{figure*}[t]
\centering
\includegraphics[width=\linewidth]{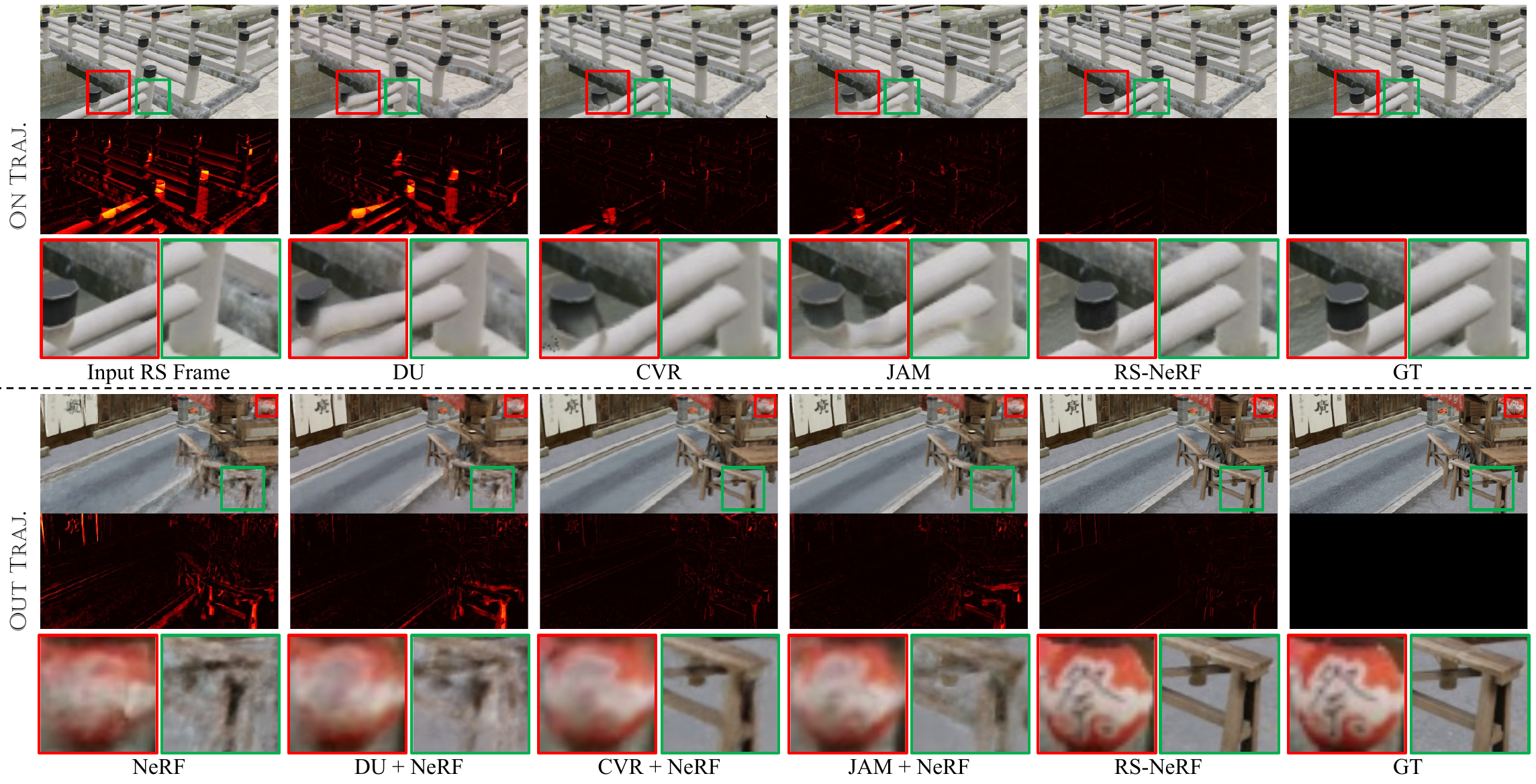}
\caption{\textbf{Qualitative comparison results on synthetic scenes.} The corresponding error maps and zoomed regions are visualized in the bottom. DU, CVR, and JAM represent DeepUnroll~\cite{liu2020deep}, CVR~\cite{fan2022context}, and JAMNet~\cite{fan2023joint} respectively.}
\label{fig:comparison}
\end{figure*}

\subsection{Comparisons}

\begin{figure*}[t]
\centering
\includegraphics[width=\linewidth]{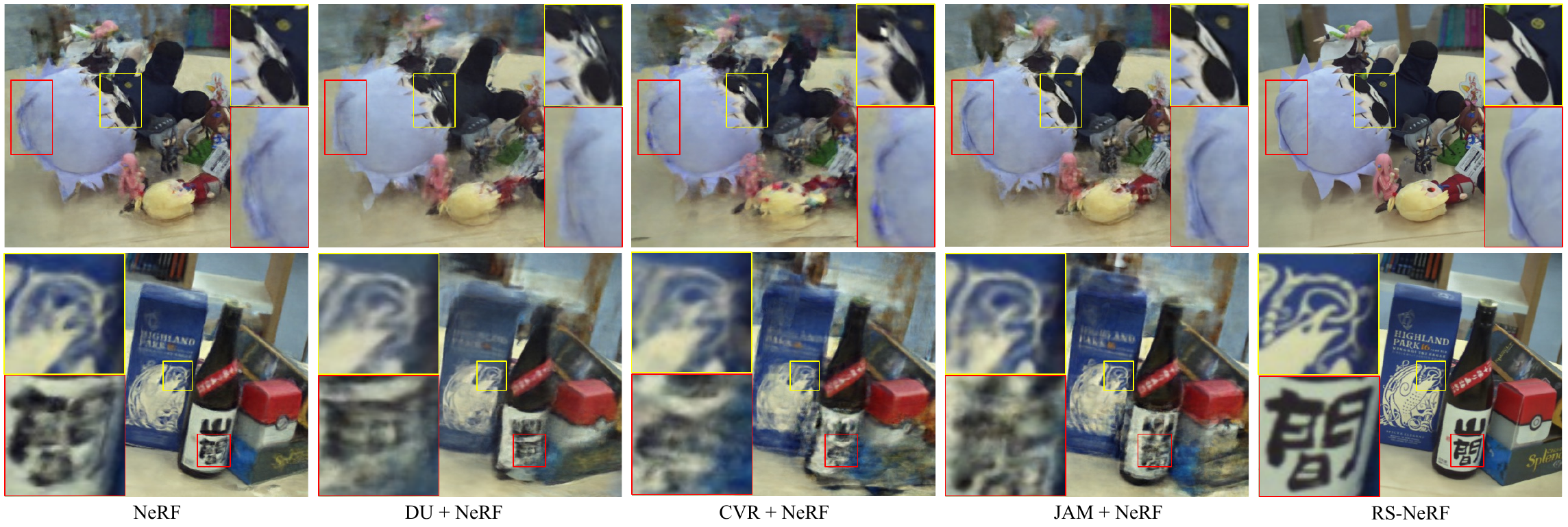}
\caption{\textbf{Qualitative comparison results on real-world scenes.} Our method achieves superior performance on real-world scenes.}
\label{fig:real}
\end{figure*}

To evaluate the effectiveness of RS correction for on-trajectory images, our method is compared with three state-of-the-art 2D RS correction techniques: DeepUnroll~\cite{liu2020deep}, CVR~\cite{fan2022context}, and JAMNet~\cite{fan2023joint}. To evaluate out-trajectory views, we selected several baselines for comparison. The most basic approach is to train NeRF directly using RS images. We further compare our method with two-stage baselines that initially apply state-of-the-art 2D RS correction methods (DeepUnroll~\cite{liu2020deep}, CVR~\cite{fan2022context}, and JAMNet~\cite{fan2023joint}) to restore GS images, followed by training NeRF with these restored images. For 2D RS correction methods, we utilized the official checkpoints for evaluation. Furthermore, we compare RS-NeRF with DRSC~\cite{qu2023fast} and USB-NeRF~\cite{li2023usb}.

\begin{table*}[t]
\centering
\resizebox{\textwidth}{!}{
\begin{tabular}{c|l|ccccccccccccccc|ccc}
\toprule
& \multirow{2}{*}{Methods} & \multicolumn{3}{c}{\normalsize{\textsc{Torii}}} & \multicolumn{3}{c}{\normalsize{\textsc{Wine}}} & \multicolumn{3}{c}{\normalsize{\textsc{Pool}}} & \multicolumn{3}{c}{\normalsize{\textsc{Factory}}} & \multicolumn{3}{c|}{\normalsize{\textsc{Tanabata}}} & \multicolumn{3}{c}{\normalsize{\textsc{Average}}} \\
& & \small{PSNR $\uparrow$} & \small{SSIM $\uparrow$} & \small{LPIPS $\downarrow$} & \small{PSNR $\uparrow$} & \small{SSIM $\uparrow$} & \small{LPIPS $\downarrow$} & \small{PSNR $\uparrow$} & \small{SSIM $\uparrow$} & \small{LPIPS $\downarrow$} & \small{PSNR $\uparrow$} & \small{SSIM $\uparrow$} & \small{LPIPS $\downarrow$} & \small{PSNR $\uparrow$} & \small{SSIM $\uparrow$} & \small{LPIPS $\downarrow$} & \small{PSNR $\uparrow$} & \small{SSIM $\uparrow$} & \small{LPIPS $\downarrow$} \\
\midrule
\multirow{3}{*}{\rotatebox[origin=c]{90}{\textsc{\tiny{On Traj.}}}}
& \textit{USB-NeRF~\cite{li2023usb}} & \secondscore{26.32} & \secondscore{.8283} & .0825 & \secondscore{24.72} & \secondscore{.7105} & .1673 & \secondscore{26.72} & \secondscore{.7130} & .1592 & \secondscore{24.32} & \secondscore{.7213} & .1369 & \secondscore{23.05} & .7359 & .1333 & \secondscore{25.02} & \secondscore{.7418} & .1359 \\
& \textit{DRSC~\cite{qu2023fast}} & 23.04 & .8008 & \secondscore{.0539} & 21.84 & .6895 & \secondscore{.0880} & 22.85 & .6962 & \secondscore{.0874} & 20.85 & .6901 & \secondscore{.0967} & 21.19 & \secondscore{.7492} & \secondscore{.0731} & 21.95 & .7251 & \secondscore{.0798} \\
& \textit{RS-NeRF} & \bestscore{30.97} & \bestscore{.9058} & \bestscore{.0395} & \bestscore{28.36} & \bestscore{.8482} & \bestscore{.0675} & \bestscore{28.50} & \bestscore{.8137} & \bestscore{.0562} & \bestscore{25.43} & \bestscore{.7673} & \bestscore{.0867} & \bestscore{26.40} & \bestscore{.8614} & \bestscore{.0651} & \bestscore{27.93} & \bestscore{.8393} & \bestscore{.0630} \\
\midrule
\multirow{3}{*}{\rotatebox[origin=c]{90}{\textsc{\tiny{Out Traj.}}}}
& \textit{USB-NeRF~\cite{li2023usb}} & \secondscore{28.95} & \secondscore{.8932} & \secondscore{.0736} & \secondscore{25.75} & \secondscore{.7446} & \secondscore{.1619} & \secondscore{28.13} & \secondscore{.7765} & \secondscore{.1576} & \secondscore{25.25} & \secondscore{.7676} & \secondscore{.1427} & \secondscore{24.62} & \secondscore{.7853} & \secondscore{.1204} & \secondscore{26.54} & \secondscore{.7935} & \secondscore{.1313} \\
& \textit{DRSC~\cite{qu2023fast}+NeRF} & 21.51 & .6790 & .2831 & 20.30 & .5145 & .3001 & 22.66 & .5614 & .3021 & 18.99 & .5295 & .2904 & 18.35 & .5643 & .2962 & 20.36 & .5697 & .2944 \\
& \textit{RS-NeRF} & \bestscore{33.07} & \bestscore{.9499} & \bestscore{.0243} & \bestscore{29.27} & \bestscore{.8703} & \bestscore{.0560} & \bestscore{28.77} & \bestscore{.8313} & \bestscore{.0458} & \bestscore{25.11} & \bestscore{.7738} & \bestscore{.0807} & \bestscore{27.02} & \bestscore{.8881} & \bestscore{.0498} & \bestscore{28.65} & \bestscore{.8627} & \bestscore{.0513} \\
\bottomrule
\end{tabular}%
}
\setlength{\abovecaptionskip}{2mm}
\caption{\textbf{Quantitative comparisons on synthetic scenes} with DRSC~\cite{qu2023fast}+NeRF and USB-NeRF~\cite{li2023usb}. We color code each result as \bestscore{best} and \secondscore{second best}.}
\label{tab:usb_tab}
\end{table*}

\begin{figure*}[t]
\includegraphics[width=\textwidth]{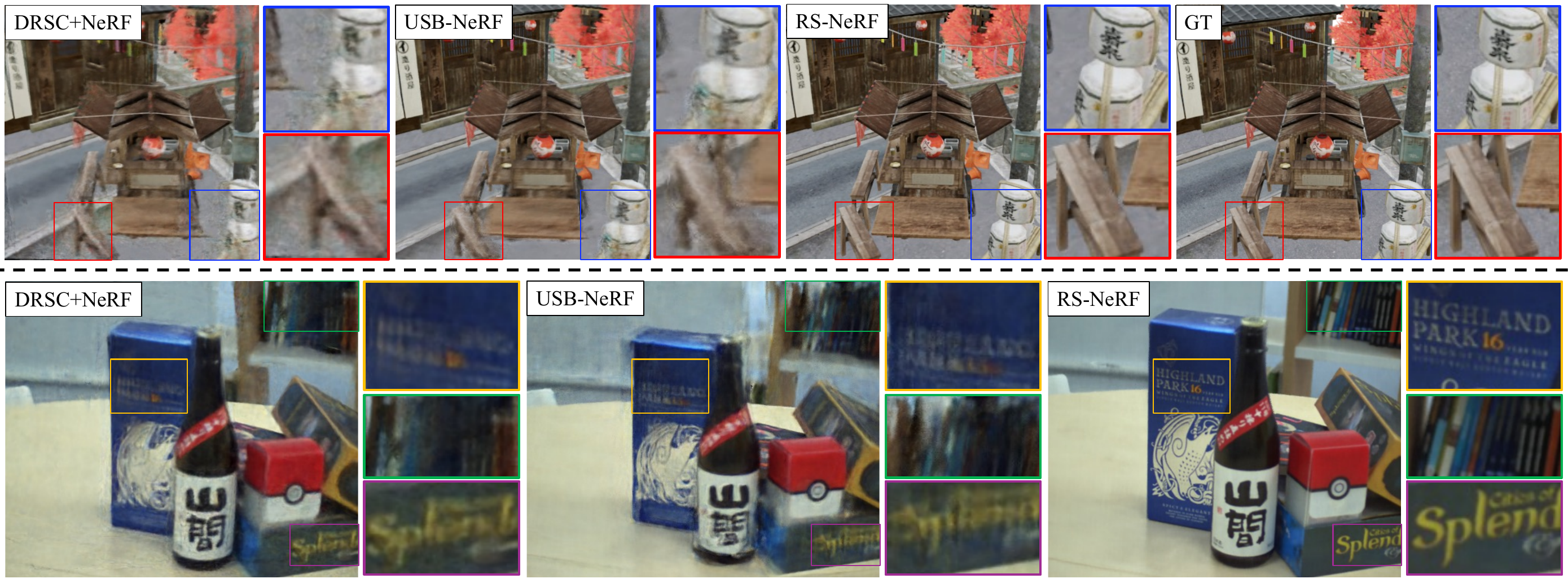}
\caption{
\textbf{Qualitative comparison results on synthetic and real-world scenes} with DRSC~\cite{qu2023fast}+NeRF and USB-NeRF~\cite{li2023usb}.
}
\label{fig:usb_fig}
\end{figure*}

\noindent \textbf{Quantitative Results.} Quantitative comparison results are shown in Tab.~\ref{tab:comparison} and Tab.~\ref{tab:usb_tab}. 2D RS correction methods do not yield satisfactory outcomes and receive obvious lower metrics due to twist artifacts and spatial misalignment. In contrast, our approach significantly exceeds these methods in all metrics from both on-trajectory and out-trajectory views by incorporating the RS formation process into the modeling and utilizing the multi-view information from continuous RS frames. USB-NeRF~\cite{li2023usb} integrates basic camera motion modeling, \ie, assigning unique camera poses for different rows to account for the RS effect, without additional technical improvements. The comparison shows that this approach fails to deliver satisfactory results compared to RS-NeRF.

\noindent \textbf{Qualitative Results.} The qualitative performance of our method is compared to others through tests on synthetic and real-world datasets. In synthetic scenes, as shown in Fig.~\ref{fig:comparison}, 2D RS correction methods struggle with twisting artifacts and spatial misalignment for on-trajectory views. In contrast, our method takes advantage of multi-view information for RS correction, leading to significantly better results. For out-trajectory views, while 2D RS correction methods combined with NeRF moderately enhance the reconstruction results, they still exhibit noticeable visual flaws, such as blurriness. RS-NeRF excels by integrating the RS formation process into the model and accurately learning the underlying 3D representation for joint RS correction and novel view synthesis.
Qualitative results in real world scenes are shown in Fig.~\ref{fig:real}. We can observe that baseline methods show obvious artifacts, particularly around object edges. Also, the performance of 2D RS correction combined with NeRF is poorer in real-world scenes than in synthetic ones, which further demonstrates the generalization issues of 2D RS correction methods with real-world data. RS-NeRF, on the other hand, significantly outperforms these methods with clear edges and detailed rendering, consistent with results in synthetic scenes. Fig.~\ref{fig:usb_fig} also demonstrates the superiority of RS-NeRF over DRSC~\cite{qu2023fast}+NeRF and USB-NeRF~\cite{li2023usb} on synthetic and real-world scenes.

{
\sidecaptionvpos{table}{t}
\begin{SCtable}[][t]
\setlength{\abovecaptionskip}{-10mm}
\resizebox{0.6\linewidth}{!}{%
\begin{tabular}{l|ccc|ccc|c}
\toprule
\multirow{2}{*}{Perturbation} & \multicolumn{3}{c|}{\textsc{On Traj.}} & \multicolumn{3}{c|}{\textsc{Out Traj.}} & \multirow{2}{*}{MSTE $\downarrow$} \\
 & \small{PSNR $\uparrow$} & \small{SSIM $\uparrow$} & \small{LPIPS $\downarrow$} & \small{PSNR $\uparrow$} & \small{SSIM $\uparrow$} & \small{LPIPS $\downarrow$} \\
\midrule
\textit{None} & 27.93 & .8393 & .0630 & 28.65 & .8627 & .0513 & .0319 \\
\textit{$U[-0.01, 0.01]$} & 27.97 & .8403 & .0610 & 28.84 & .8680 & .0485 & .0297 \\
\textit{$U[-0.03, 0.03]$} & 28.31 & .8479 & .0578 & 29.00 & .8693 & .0470 & .0304 \\
\textit{$U[-0.05, 0.05]$} & 28.47 & .8530 & .0580 & 29.24 & .8769 & .0466 & .0308 \\
\textit{$U[-0.07, 0.07]$} & 28.43 & .8504 & .0597 & 29.25 & .8771 & .0471 & .0336 \\
\bottomrule
\end{tabular}%
}
\caption{\textbf{Quantitative results on synthetic dataset for different noise perturbations on pose initialization.} The results demonstrate that our model is robust to the accuracy of pose initialization.}
\label{tab:noise}
\end{SCtable}
}

\subsection{Discussion and Limitations}

\noindent \textbf{Robustness to pose initialization.} We evaluated the robustness of our method to the precision of pose initialization on synthetic dataset by introducing random noise of varying scales to the initial poses in the $\textbf{SE}(3)$ space and using these altered poses for initialization. The results, as detailed in Tab.~\ref{tab:noise}, indicate that our method maintains superior performance across different levels of noise perturbation. 
These results confirm the robustness of RS-NeRF in the accuracy of pose initialization. However, it is important to note that our method might fail when completely random poses are used for initialization. This challenge is also an ongoing issue in the field of NeRF-without-Poses, as discussed in works like~\cite{bian2023nope,lin2021barf}. Addressing this limitation is a direction that we plan to pursue in future research.

\noindent \textbf{Limitations.} Similar to previous works that address NeRF with abnormal inputs~\cite{ma2022deblur,wang2023bad}, our model is designed to handle degradation (referred to the RS effects in our context) caused by camera movements, since we primarily model the camera trajectory. When encountering scenarios where RS effects are caused by object movements, \ie, dynamic scenes, our method may fail and produce unsatisfying results. We plan to involve the modeling of object movements based on dynamic NeRFs in future work.

\section{Conclusion}
\label{sec:conclusion}

In this paper, we present RS-NeRF, a method designed to synthesize normal images from novel views using input with Rolling Shutter (RS) distortions. We formulate a physical model that replicates the image formation process under RS conditions and jointly optimizes the NeRF parameters and the camera extrinsic for each image row. We then identify and address the fundamental limitations of the standard RS-NeRF model by introducing two innovative approaches: trajectory smoothness regularization and the multi-sampling algorithm. Through rigorous experimentation, we demonstrate that RS-NeRF surpasses previous methods in both synthetic and real-world scenarios, confirming its efficiency in rectifying distortions caused by RS cameras.


%
%
\bibliographystyle{splncs04}
\bibliography{main}

\begin{thebibliography}{10}
\providecommand{\url}[1]{\texttt{#1}}
\providecommand{\urlprefix}{URL }
\providecommand{\doi}[1]{https://doi.org/#1}

\bibitem{albl2019rolling}
Albl, C., Kukelova, Z., Larsson, V., Pajdla, T.: Rolling shutter camera absolute pose. IEEE transactions on pattern analysis and machine intelligence  \textbf{42}(6),  1439--1452 (2019)

\bibitem{attal2021torf}
Attal, B., Laidlaw, E., Gokaslan, A., Kim, C., Richardt, C., Tompkin, J., O'Toole, M.: T{\"o}rf: Time-of-flight radiance fields for dynamic scene view synthesis. Advances in neural information processing systems  \textbf{34},  26289--26301 (2021)

\bibitem{barron2021mip}
Barron, J.T., Mildenhall, B., Tancik, M., Hedman, P., Martin-Brualla, R., Srinivasan, P.P.: Mip-nerf: A multiscale representation for anti-aliasing neural radiance fields. In: Proceedings of the IEEE/CVF International Conference on Computer Vision. pp. 5855--5864 (2021)

\bibitem{barron2022mip}
Barron, J.T., Mildenhall, B., Verbin, D., Srinivasan, P.P., Hedman, P.: Mip-nerf 360: Unbounded anti-aliased neural radiance fields. In: Proceedings of the IEEE/CVF Conference on Computer Vision and Pattern Recognition. pp. 5470--5479 (2022)

\bibitem{bian2023nope}
Bian, W., Wang, Z., Li, K., Bian, J.W., Prisacariu, V.A.: Nope-nerf: Optimising neural radiance field with no pose prior. In: Proceedings of the IEEE/CVF Conference on Computer Vision and Pattern Recognition. pp. 4160--4169 (2023)

\bibitem{cao2022learning}
Cao, M., Zhong, Z., Wang, J., Zheng, Y., Yang, Y.: Learning adaptive warping for real-world rolling shutter correction. In: Proceedings of the IEEE/CVF Conference on Computer Vision and Pattern Recognition. pp. 17785--17793 (2022)

\bibitem{chen2022aug}
Chen, T., Wang, P., Fan, Z., Wang, Z.: Aug-nerf: Training stronger neural radiance fields with triple-level physically-grounded augmentations. In: Proceedings of the IEEE/CVF Conference on Computer Vision and Pattern Recognition. pp. 15191--15202 (2022)

\bibitem{dave2022pandora}
Dave, A., Zhao, Y., Veeraraghavan, A.: Pandora: Polarization-aided neural decomposition of radiance. In: Computer Vision--ECCV 2022: 17th European Conference, Tel Aviv, Israel, October 23--27, 2022, Proceedings, Part VII. pp. 538--556. Springer (2022)

\bibitem{fan2021sunet}
Fan, B., Dai, Y., He, M.: Sunet: symmetric undistortion network for rolling shutter correction. In: Proceedings of the IEEE/CVF International Conference on Computer Vision. pp. 4541--4550 (2021)

\bibitem{fan2022context}
Fan, B., Dai, Y., Zhang, Z., Liu, Q., He, M.: Context-aware video reconstruction for rolling shutter cameras. In: Proceedings of the IEEE/CVF Conference on Computer Vision and Pattern Recognition. pp. 17572--17582 (2022)

\bibitem{fan2023joint}
Fan, B., Mao, Y., Dai, Y., Wan, Z., Liu, Q.: Joint appearance and motion learning for efficient rolling shutter correction. In: Proceedings of the IEEE/CVF Conference on Computer Vision and Pattern Recognition. pp. 5671--5681 (2023)

\bibitem{garbin2021fastnerf}
Garbin, S.J., Kowalski, M., Johnson, M., Shotton, J., Valentin, J.: Fastnerf: High-fidelity neural rendering at 200fps. In: Proceedings of the IEEE/CVF International Conference on Computer Vision. pp. 14346--14355 (2021)

\bibitem{guo2022nerfren}
Guo, Y.C., Kang, D., Bao, L., He, Y., Zhang, S.H.: Nerfren: Neural radiance fields with reflections. In: Proceedings of the IEEE/CVF Conference on Computer Vision and Pattern Recognition. pp. 18409--18418 (2022)

\bibitem{hedborg2012rolling}
Hedborg, J., Forss{\'e}n, P.E., Felsberg, M., Ringaby, E.: Rolling shutter bundle adjustment. In: 2012 IEEE Conference on Computer Vision and Pattern Recognition. pp. 1434--1441. IEEE (2012)

\bibitem{kim2017rrd}
Kim, J.H., Latif, Y., Reid, I.: Rrd-slam: Radial-distorted rolling-shutter direct slam. In: 2017 IEEE International Conference on Robotics and Automation (ICRA). pp. 5148--5154. IEEE (2017)

\bibitem{kingma2014adam}
Kingma, D.P., Ba, J.: Adam: A method for stochastic optimization. arXiv preprint arXiv:1412.6980  (2014)

\bibitem{klenk2023nerf}
Klenk, S., Koestler, L., Scaramuzza, D., Cremers, D.: E-nerf: Neural radiance fields from a moving event camera. IEEE Robotics and Automation Letters  (2023)

\bibitem{klingner2013street}
Klingner, B., Martin, D., Roseborough, J.: Street view motion-from-structure-from-motion. In: Proceedings of the IEEE International Conference on Computer Vision. pp. 953--960 (2013)

\bibitem{li2023usb}
Li, M., Wang, P., Zhao, L., Liao, B., Liu, P.: {USB}-ne{RF}: Unrolling shutter bundle adjusted neural radiance fields. In: The Twelfth International Conference on Learning Representations (2024), \url{https://openreview.net/forum?id=igfDXfMvm5}

\bibitem{lin2021barf}
Lin, C.H., Ma, W.C., Torralba, A., Lucey, S.: Barf: Bundle-adjusting neural radiance fields. In: IEEE International Conference on Computer Vision ({ICCV}) (2021)

\bibitem{liu2020neural}
Liu, L., Gu, J., Zaw~Lin, K., Chua, T.S., Theobalt, C.: Neural sparse voxel fields. Advances in Neural Information Processing Systems  \textbf{33},  15651--15663 (2020)

\bibitem{liu2020deep}
Liu, P., Cui, Z., Larsson, V., Pollefeys, M.: Deep shutter unrolling network. In: Proceedings of the IEEE/CVF Conference on Computer Vision and Pattern Recognition. pp. 5941--5949 (2020)

\bibitem{liu2021mba}
Liu, P., Zuo, X., Larsson, V., Pollefeys, M.: Mba-vo: Motion blur aware visual odometry. In: Proceedings of the IEEE/CVF International Conference on Computer Vision. pp. 5550--5559 (2021)

\bibitem{ma2022deblur}
Ma, L., Li, X., Liao, J., Zhang, Q., Wang, X., Wang, J., Sander, P.V.: Deblur-nerf: Neural radiance fields from blurry images. In: Proceedings of the IEEE/CVF Conference on Computer Vision and Pattern Recognition. pp. 12861--12870 (2022)

\bibitem{mildenhall2021nerf}
Mildenhall, B., Srinivasan, P.P., Tancik, M., Barron, J.T., Ramamoorthi, R., Ng, R.: Nerf: Representing scenes as neural radiance fields for view synthesis. Communications of the ACM  \textbf{65}(1),  99--106 (2021)

\bibitem{qu2023fast}
Qu, D., Liao, B., Zhang, H., Ait-Aider, O., Lao, Y.: Fast rolling shutter correction in the wild. IEEE Transactions on Pattern Analysis and Machine Intelligence  (2023)

\bibitem{reiser2021kilonerf}
Reiser, C., Peng, S., Liao, Y., Geiger, A.: Kilonerf: Speeding up neural radiance fields with thousands of tiny mlps. In: Proceedings of the IEEE/CVF International Conference on Computer Vision. pp. 14335--14345 (2021)

\bibitem{rengarajan2017unrolling}
Rengarajan, V., Balaji, Y., Rajagopalan, A.: Unrolling the shutter: Cnn to correct motion distortions. In: Proceedings of the IEEE Conference on computer Vision and Pattern Recognition. pp. 2291--2299 (2017)

\bibitem{saurer2013rolling}
Saurer, O., Koser, K., Bouguet, J.Y., Pollefeys, M.: Rolling shutter stereo. In: Proceedings of the IEEE International Conference on Computer Vision. pp. 465--472 (2013)

\bibitem{saurer2016sparse}
Saurer, O., Pollefeys, M., Lee, G.H.: Sparse to dense 3d reconstruction from rolling shutter images. In: Proceedings of the IEEE Conference on Computer Vision and Pattern Recognition. pp. 3337--3345 (2016)

\bibitem{schonberger2016structure}
Schonberger, J.L., Frahm, J.M.: Structure-from-motion revisited. In: Proceedings of the IEEE conference on computer vision and pattern recognition. pp. 4104--4113 (2016)

\bibitem{schonberger2016pixelwise}
Sch{\"o}nberger, J.L., Zheng, E., Frahm, J.M., Pollefeys, M.: Pixelwise view selection for unstructured multi-view stereo. In: Computer Vision--ECCV 2016: 14th European Conference, Amsterdam, The Netherlands, October 11-14, 2016, Proceedings, Part III 14. pp. 501--518. Springer (2016)

\bibitem{teed2020raft}
Teed, Z., Deng, J.: Raft: Recurrent all-pairs field transforms for optical flow. In: Computer Vision--ECCV 2020: 16th European Conference, Glasgow, UK, August 23--28, 2020, Proceedings, Part II 16. pp. 402--419. Springer (2020)

\bibitem{vasu2018occlusion}
Vasu, S., Rajagopalan, A., et~al.: Occlusion-aware rolling shutter rectification of 3d scenes. In: Proceedings of the IEEE Conference on Computer Vision and Pattern Recognition. pp. 636--645 (2018)

\bibitem{verbin2022refnerf}
Verbin, D., Hedman, P., Mildenhall, B., Zickler, T., Barron, J.T., Srinivasan, P.P.: {Ref-NeRF}: Structured view-dependent appearance for neural radiance fields. CVPR  (2022)

\bibitem{wang2023bad}
Wang, P., Zhao, L., Ma, R., Liu, P.: Bad-nerf: Bundle adjusted deblur neural radiance fields. In: Proceedings of the IEEE/CVF Conference on Computer Vision and Pattern Recognition. pp. 4170--4179 (2023)

\bibitem{xu2022point}
Xu, Q., Xu, Z., Philip, J., Bi, S., Shu, Z., Sunkavalli, K., Neumann, U.: Point-nerf: Point-based neural radiance fields. In: Proceedings of the IEEE/CVF Conference on Computer Vision and Pattern Recognition. pp. 5438--5448 (2022)

\bibitem{zhang2018unreasonable}
Zhang, R., Isola, P., Efros, A.A., Shechtman, E., Wang, O.: The unreasonable effectiveness of deep features as a perceptual metric. In: Proceedings of the IEEE conference on computer vision and pattern recognition. pp. 586--595 (2018)

\bibitem{zhong2022bringing}
Zhong, Z., Cao, M., Sun, X., Wu, Z., Zhou, Z., Zheng, Y., Lin, S., Sato, I.: Bringing rolling shutter images alive with dual reversed distortion. In: European Conference on Computer Vision. pp. 233--249. Springer (2022)

\bibitem{zhong2021towards}
Zhong, Z., Zheng, Y., Sato, I.: Towards rolling shutter correction and deblurring in dynamic scenes. In: Proceedings of the IEEE/CVF Conference on Computer Vision and Pattern Recognition. pp. 9219--9228 (2021)

\bibitem{zhou2022evunroll}
Zhou, X., Duan, P., Ma, Y., Shi, B.: Evunroll: Neuromorphic events based rolling shutter image correction. In: Proceedings of the IEEE/CVF Conference on Computer Vision and Pattern Recognition. pp. 17775--17784 (2022)

\bibitem{zhuang2019learning}
Zhuang, B., Tran, Q.H., Ji, P., Cheong, L.F., Chandraker, M.: Learning structure-and-motion-aware rolling shutter correction. In: Proceedings of the IEEE/CVF Conference on Computer Vision and Pattern Recognition. pp. 4551--4560 (2019)

\end{thebibliography}
\end{document}